\pdfoutput=1

\documentclass[11pt]{article}

\usepackage[]{ACL2023}

\usepackage{times}
\usepackage{latexsym}

\usepackage[T1]{fontenc}

\usepackage[utf8]{inputenc}

\usepackage{microtype}

\usepackage{inconsolata}
\usepackage{soul}
\usepackage{utfsym}
\usepackage{pifont}
\usepackage{enumitem}
\usepackage{amstext}
\usepackage{booktabs}
\usepackage{graphicx}
\usepackage{mathtools}
\usepackage{algorithmic}
\usepackage{algorithm}
\usepackage{amsfonts,amssymb}
\usepackage{multirow}
\usepackage{caption}
\usepackage{subcaption}
\usepackage{array}
\usepackage{bm}
\usepackage{makecell}

\usepackage{xcolor,colortbl}
\usepackage{CJKutf8}
\usepackage{balance}
\newcommand{\cchar}[1]{\begin{CJK*}{UTF8}{gkai}#1\end{CJK*}}
\newcommand{\ccharb}[1]{\begin{CJK*}{UTF8}{min}#1\end{CJK*}}

\usepackage{tikz}
\newcommand*{\mybox}[2]{\tikz[anchor=base,baseline=0pt,rounded corners=0pt, inner sep=0.2mm] \node[fill=#1] (X) {#2};}

\def\bcbaux#1#2 #3\endbcb{%
  \colorbox{#1}{\strut#2}%
  \ifx\relax#3\relax\def\next{}\else%
    \colorbox{#1}{ \strut}%
    \allowbreak%
    \def\next{\bcbaux{#1}#3\endbcb}%
  \fi%
  \next%
}

\title{mPMR: A Multilingual Pre-trained Machine Reader at Scale\thanks{~~This work was supported by Alibaba Group through Alibaba Research Intern Program. The work described in this paper was also partially supported by a grant from the Research Grant Council of the Hong Kong Special Administrative Region, China (Project Code: 14200719). $^\dagger$ This work was done when Weiwen Xu was an intern at Alibaba DAMO Academy. $^\ddagger$ Xin Li is the corresponding author.}}

\author{
Weiwen Xu\raisebox{4pt}{\small $12$,}$^\dagger$ \quad Xin Li\raisebox{4pt}{\small $2$,}$^\ddagger$ \quad \textbf{Wai Lam}\raisebox{4pt}{\small $1$} \quad \textbf{Lidong Bing}\raisebox{4pt}{\small $2$}\\
\raisebox{4pt}{\small $1$}The Chinese University of Hong Kong \\
\raisebox{4pt}{\small $2$}DAMO Academy, Alibaba Group \\
{\tt \{wwxu,wlam\}@se.cuhk.edu.hk} \quad 
{\tt \{xinting.lx,l.bing\}@alibaba-inc.com}
}

\date{}

\begin{document}
\maketitle

\begin{abstract}
We present multilingual Pre-trained Machine Reader (mPMR), a novel method for multilingual machine reading comprehension (MRC)-style pre-training.
mPMR aims to guide  multilingual pre-trained language models (mPLMs) to perform natural language understanding (NLU) including both sequence classification and span extraction in multiple languages.
To achieve cross-lingual generalization when only source-language fine-tuning data is available, existing mPLMs solely transfer NLU capability from a source language to target languages.
In contrast, mPMR allows the direct inheritance of multilingual NLU capability from the MRC-style pre-training to downstream tasks.
Therefore, mPMR acquires better NLU capability for target languages.
mPMR also provides a unified solver for tackling cross-lingual span extraction and sequence classification, thereby enabling the extraction of rationales to explain the sentence-pair classification process.\footnote{The code, data, and checkpoints are released at \url{https://github.com/DAMO-NLP-SG/PMR}}

\end{abstract}
\section{Introduction}
Multilingual pre-trained language models, acro\-nymed as mPLMs, have demonstrated strong Natural language understanding (NLU) capability in a  wide range of languages \cite{xue-etal-2021-mt5,cai-etal-2021-multilingual-amr,cai-etal-2022-retrofitting,conneau-etal-2020-unsupervised,ding-etal-2022-globalwoz,DBLP:conf/ijcai/LiHYNB020}.
In particular, mPLMs can maintain exceptional cross-lingual language understanding (XLU) capability on unseen \textit{target} languages though mPLMs are only fine-tuned on resource-rich \textit{source} languages like English.

It has been proved that optimizing cross-lingual representations of mPLMs can improve XLU capability. 
For example, cross-lingual supervisions, such as parallel sentences \cite{NEURIPS2019_c04c19c2} or bilingual dictionaries \cite{conneau-etal-2020-emerging} could enhance cross-lingual representations with better language alignment.
XLM-R \cite{conneau-etal-2020-unsupervised} and mT5 \cite{xue-etal-2021-mt5} showed that appropriately incorporating more languages during pre-training leads to better cross-lingual representations.
A few works enriched the cross-lingual representations with factual knowledge through the utilization of multilingual mentions of  entities~\cite{calixto-etal-2021-wikipedia, ri-etal-2022-mluke} and relations~\cite{liu-etal-2022-enhancing-multilingual,jiang2022xlm} annotated in knowledge graphs.
Despite their differences, the above methods essentially constructed more diverse multilingual corpora for pre-training mPLMs.
These mPLMs would presumably meet their saturation points and are known to suffer from \textit{curse of multilinguality} \cite{conneau-etal-2020-unsupervised,pfeiffer-etal-2022-lifting,berend-2022-combating}.
Under this situation, introducing more training data from either existing \cite{pfeiffer-etal-2022-lifting} or unseen \cite{conneau-etal-2020-unsupervised} languages for enhancing mPLMs may not bring further improvement or even be detrimental to their cross-lingual representations. 

\begin{figure}[t]
    \centering
    \includegraphics[scale=0.38]{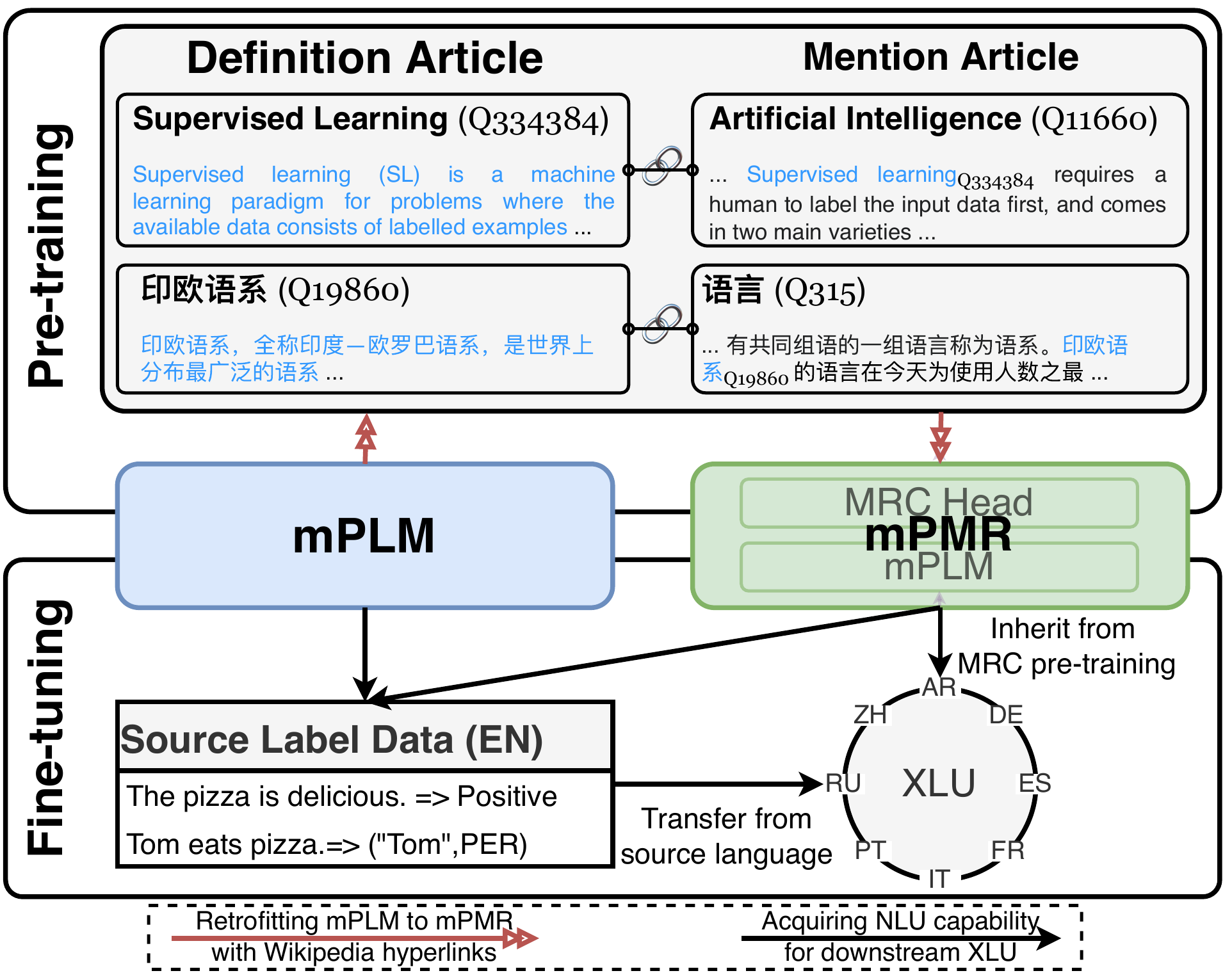}
    \caption{Pre-training and fine-tuning  of mPMR. }
    \label{fig:intro}
\end{figure}

In the paper, instead of training a new mPLM with better cross-lingual representations, we propose \textbf{m}ultilingual \textbf{P}re-trained \textbf{M}achine \textbf{R}eader (mPMR) to directly guide existing mPLMs to perform NLU in various languages.  As shown in Figure~\ref{fig:intro},  mPMR resembles PMR \cite{xu2022clozing} for constructing multilingual machine reading comprehension (MRC)-style data with Wikipedia hyperlinks.
These data are used to retrofit an mPLM into an mPMR through an MRC-style continual pre-training.
During retrofitting process (i.e., pre-training), mPMR jointly learns the general sequence classification and span extraction capability for multiple languages.
In XLU fine-tuning, mPLMs solely rely on cross-lingual representations to transfer NLU capability from a source language to target languages. 
By contrast, mPMR enables the direct inheritance of multilingual NLU capability from the MRC-style pre-training to downstream tasks in a unified MRC formulation, which alleviates the discrepancies between source-language fine-tuning and target-language inference \cite{zhou2022enhancing,zhou2022conner,zhou-etal-2023-improving}. Therefore, mPMR shows greater potential in XLU than mPLMs.

To improve the scalability of mPMR across multiple languages,  we further propose \textit{Unified Q/C Construction} and \textit{Stochastic answer position} strategies for refining the curation of MRC data.
With these two strategies, mPMR can better generalize to low-resource languages and becomes more robust to position bias \cite{ko-etal-2020-look}.

The experimental results show that mPMR obtains clear improvements over XLM-R \cite{conneau-etal-2020-unsupervised} on span extraction, with an average improvement of up to 12.6 F1 on TyDiQA, and 8.7 F1 on WikiAnn respectively.
The analysis reveals that mPMR benefits from more multilingual MRC data for pre-training.
We also found that mPMR converges faster in downstream tasks and is capable of using its strong extraction capability for explaining the sequence classification process.

\section{mPMR}
We present the MRC model and training data of mPMR. We closely follow PMR~\cite{xu2022clozing} and introduce the modifications for enabling multilingual MRC-style pre-training.

\begin{table*}[]
    \centering
    \small
    \setlength{\tabcolsep}{1mm}{
    \begin{tabular}{lcccccccccc} 
    \toprule
    \multirow{2}{*}{\textbf{Model}} & \multirow{2}{*}{\textbf{\#Params}} & \multicolumn{3}{c}{ \textbf{EQA}} & \multicolumn{2}{c}{ \textbf{NER}} & \textbf{ABSA} & \multicolumn{2}{c}{ \textbf{Sentence Pair}} & \multirow{2}{*}{\textbf{Avg.}}\\
     \cmidrule(lr){3-5}\cmidrule(lr){6-7}\cmidrule(lr){8-8}\cmidrule(lr){9-10}
    & & \textbf{XQuAD} & \textbf{MLQA} & \textbf{TyDiQA} & \textbf{WikiAnn} & \textbf{CoNLL} & \textbf{SemEval16} & \textbf{PAWS-X} & \textbf{XNLI}& \\ \midrule
    Metrics &       & F1 / EM       & F1 / EM       & F1 / EM       & F1    & F1    & F1    & Acc.  & Acc.    \\ \midrule
    XLM-R       &550M   &76.6 / 60.8    & 71.6 / 53.2   & 65.1 / 45.0   & 65.4  & 82.0  & 66.9$^{\ddag}$  & 86.4 & 79.2 & 74.2    \\
    mT5         & 580M  & 67.0 / 49.0   & 64.6 / 45.0   & 57.2 / 41.2   & 55.7  & \;71.0$^{\ddag}$    & 62.5$^{\ddag}$    & 86.4 & 75.4 & 67.5 \\
    VECO        & 550M  & 77.3 / 61.8   & 71.7 / 53.2   & 67.6 / 49.1   & 65.7  & \;81.3$^{\ddag}$    & 63.0$^{\ddag}$    & \textbf{88.7} & \textbf{79.9} & 74.4  \\
    mLUKE-W     & 561M  & \textbf{79.6} /\;\;\;-\;\;\; & 72.7 /\;\;\;-\;\;\;  & \;65.2 / 48.5$^{\ddag}$ &  \;67.7$^{\ddag}$   & 83.0  & 61.2$^{\ddag}$  & \;88.2$^{\ddag}$   & \;79.4$^{\ddag}$ & 74.6 \\
    Wiki-CL     & 550M  &72.1 / 56.9    & 70.8 / 50.5   & 73.2 / 57.3   & 64.7  & -     & -     & 88.4  & 79.2  & - \\
    KMLM        & 550M  & 77.3 / 61.7   & 72.1 / 53.7   & 67.9 / 50.4   & \;66.7$^{\ddag}$  & 83.2  & 66.1$^{\ddag}$  & 88.0  & 79.2 & 75.1    \\
    \midrule
\multicolumn{11}{c}{\emph{Our MRC Formulation}} \\
\midrule
    XLM-R$_{\tt base}$  & 270M  & 70.8 / 56.9   & 64.4 / 47.9   & 50.8 / 38.2   & 57.9  & 79.2  & 60.0  & 85.0 & 73.3 & 67.7\\
    mPMR$_{\tt base}$   & 270M  & 74.0 / 59.5   & 65.3 / 48.7   & 63.4 / 49.0   & 66.6  & 81.7  & 62.1  & 86.1  & 73.6  & 71.6\\
    XLM-R   & 550M  & 77.1 / 61.3   & 71.5 / 53.9   & 67.4 / 51.6   & 63.6  & 81.4  & 66.1  &  86.9 & 78.6  & 74.1   \\
    mPMR    & 550M  & 79.2 / \textbf{64.4}   & \textbf{73.1 / 55.4}   & \textbf{74.7 / 58.3}   & \textbf{70.7}  & \textbf{84.1}  & \textbf{68.2}  & 88.0  & 79.3    & \textbf{77.2}\\
    \bottomrule
    \end{tabular}}
    \caption{The results of all XLU tasks. We report the average results of all languages for each dataset. We also compute the overall average score among all datasets in the \textbf{Avg.} column. We reproduce the missing results with the $^{\ddag}$ label. Some results of Wiki-CL are left blank because they do not release their model checkpoint.}
    \label{tab:XLU}
\end{table*}

\subsection{Model Pre-training}

Our mPMR follows the same MRC architecture of \citet{xu2022clozing,xu2022peerda} with an encoder and an extractor.
The encoder maps input tokens $X$, the concatenation of the query $Q$, the context $C$, and special markers (i.e., ${\tt [CLS]}$ and ${\tt [SEP]}$), into hidden representations $H$. For any two tokens $X_i$ and $X_j$ ($i < j$), the extractor receives their contextualized representations $H_i$ and $H_j$ and predicts the probability score $S_{i,j}$ indicating the probability of the token span $X_{i:j}$ being the answer to the query $Q$.

mPMR is guided with the Wiki Anchor Extraction (WAE) objective to train both the encoder and the extractor.
WAE checks if the answer to the query exists in the context. If so, WAE would first regard the query and the context to be relevant and extracts the ${\tt [CLS]}$ token as a sequence-level relevance indicator. WAE would then extract all corresponding answers from the context. 

\subsection{Multilingual MRC Data}
\label{sec:data}
Training mPMR requires the existence of labeled (query, context, answer) triplets.
To obtain such data, we collected Wikipedia articles with anchor annotations for 24 languages, which are the most widely used and cover a reasonable number of languages used in XLU tasks \cite{ri-etal-2022-mluke}. 

As shown in Figure \ref{fig:intro}, we utilized a Wikipedia anchor to obtain a pair of correlated articles. One side of the pair is the article that provides in-depth descriptions of the anchor entity, which we defined as the \textit{definition article}. The other side of the pair is named as the \textit{mention article}, which mentions the specific anchor text\footnote{definition/mention article refers to home/reference article of \citet{xu2022clozing}.}. 
We composed an answerable MRC example in which the anchor is the answer, the surrounding text of the anchor in the mention article is the context, and the definition of the anchor entity in the definition article is the query.
Additionally, we can generate an unanswerable MRC example by pairing a query with an irrelevant context without anchor association. 

\paragraph{Unified Q/C Construction.}
PMR constructed the MRC query and context as valid sentences so as to keep the text coherent.
However, sentence segmentation tools are usually not available for low-resource languages.
To remedy this, we did not apply sentence segmentation but only preprocess Wikipedia articles with word tokenization in mPMR.
For each anchor, the MRC query comprises the first $Q$ words in the definition article.
To prevent information leakage during pre-training, similar to PMR, we anonymized the anchor entity in the query to the \texttt{[MASK]} token.
The MRC context consists of $C$ words surrounding the anchor.

\paragraph{Stochastic Answer Position.}
As mentioned by \citet{ko-etal-2020-look}, the model is prone to overfitting to the position shortcut if the answer in the context exhibits a fixed position pattern. In our case, suppose that the MRC context consists of $C/2$ words on both the left and right sides of the anchor, the model may learn the shortcut that the middle part of the context is likely to be the answer.
To prevent such position bias, we propose a stochastic answer position method, which allows the answer to be presented in any position within the context.
Specifically, given an anchor in a Wikipedia article, the context comprises $\xi$ words preceding the anchor and the $C -\xi$ words following the anchor, where $\xi$ is a random integer ranging from 0 to $C$ and varies across different contexts.
In accordance with PMR, we treated all text spans identical to the anchor in the current context as valid answers.


\section{Experimental Setup}
\paragraph{Implementation Details.}
In mPMR, the encoder is loaded from XLM-R \cite{conneau-etal-2020-unsupervised} and the extractor is randomly initialized. 
Both components are then continually pre-trained using the multilingual MRC data that we constructed.
More hyper-parameters can be found in Appendix~\ref{app:impl}.

\paragraph{Downstream XLU Tasks.}
 We evaluated mPMR on a series of span extraction tasks, including Extractive Question Answering (EQA), Named Entity Recognition (NER), and Aspect-Based Sentiment Analysis (ABSA).
We also evaluated our mPMR on two sequence classification tasks.
We followed \citet{xu2022clozing} to convert all tasks into MRC formulation to effectively leverage the knowledge that is acquired during MRC-style pre-training.
For EQA, we used XQuAD \cite{artetxe-etal-2020-cross}, MLQA \cite{lewis-etal-2020-mlqa}, and TyDiQA \cite{clark-etal-2020-tydi}.
For NER, we used WikiAnn \cite{pan-etal-2017-cross} and CoNLL \cite{tjong-kim-sang-2002-introduction,tjong-kim-sang-de-meulder-2003-introduction}.
SemEval16 \cite{pontiki-etal-2016-semeval} was used for ABSA task.
Regarding the sequence classification, we used XNLI \cite{conneau-etal-2018-xnli} and PAWS-X \cite{yang-etal-2019-paws}. Additional dataset information and concrete examples are provided in Appendix~\ref{app:dataset}

\paragraph{Baselines.}
We compared mPMR with recent methods on improving cross-lingual representations, including 1) models pre-trained on a large number of languages: XLM-R \cite{conneau-etal-2020-unsupervised}, mT5 \cite{xue-etal-2021-mt5}, and VECO \cite{luo-etal-2021-veco}; 2) models that exploited multilingual entity information: Wiki-CL \cite{calixto-etal-2021-wikipedia}, and mLUKE-W \cite{ri-etal-2022-mluke}; and 3) Model that utilized multilingual relation information: KMLM \cite{liu-etal-2022-enhancing-multilingual}.
For a fair comparison, all models have approximately the same parameter size.
\begin{table*}[]
    \centering
    \small
    \setlength{\tabcolsep}{1.5mm}{
    \begin{tabular}{c|l|c|cccc} 
    \toprule
    \textbf{Index} &  \textbf{Model} & \textbf{\#Lang} & \textbf{PAWS-X} & \textbf{XQuAD} & \textbf{WikiAnn} & \textbf{Avg.}  \\ 
        \midrule
    \#1 & XLM-R$_{\tt base}$        &0  & 85.0  & 70.8  & 57.9 & 71.2\\ 
    \#2 &     \#1 + MRC data in English      & 1 & 85.2 (0.2$\uparrow$) & 71.0 (0.2$\uparrow$) & 59.5 (1.6$\uparrow$)  & 71.9 (0.7$\uparrow$)\\
    \#3 &    \#2 + Stochastic Answer Position   & 1 & 85.5 (0.3$\uparrow$) & 73.0 (2.0$\uparrow$)  & 60.0 (0.5$\uparrow$)  & 72.8 (0.9$\uparrow$)\\ 
    \#4 &    \#3 + MRC data in more languages   & 10 & 85.9 (0.4$\uparrow$) & 73.5 (0.5$\uparrow$) & 64.7 (4.7$\uparrow$) & 74.7 (1.9$\uparrow$)\\
    \#5 &   \#4 + MRC data in even more languages (mPMR$_{\tt base}$) & 24  & \textbf{86.1} (0.2$\uparrow$) & \textbf{74.0} (0.5$\uparrow$) & \textbf{66.6} (1.9$\uparrow$)  & \textbf{75.6} (0.9$\uparrow$) \\ \bottomrule
    \end{tabular}}
    \caption{The process of retrofitting XLM-R into mPMR using multilingual MRC data (English$\rightarrow$10 languages$\rightarrow$24 languages) and our Stochastic Answer Position method. Each row accumulates modifications from all rows above.}
    \label{tab:pt}
\end{table*}

\begin{table*}[]
    \centering
    \small
    \renewcommand{\arraystretch}{1.25}
    \setlength{\tabcolsep}{1mm}{
    \begin{tabular}{l|p{68mm}|p{68mm}}
    \Xhline{3\arrayrulewidth}
          \textbf{Label} & \textbf{Sentence 1} &  \textbf{Sentence 2}\\ \hline
         Entailment  & Rami Nieminen ( born February 25 , 1966 ) \mybox{red!15}{{{is a Finnish footballer}}}.
            & Rami Nieminen ( born 25 February 1966 ) is a \mybox{red!15}{{{Finnish former footballer}}}. \\ \hline
         Contradiction  & In 1938 he became the Government Anthropologist of the \mybox{red!15}{{{Egyptian-Anglo Sudan}}} and conducted fieldwork with the Nuba.
            & In 1938 he became the government anthropologist of the \mybox{red!15}{{{anglo-Egyptian}}} Sudan and led fieldwork with the Nuba . \\ \hline
         Entailment  & \cchar{Stipsits 出生于\mybox{red!15}{{{科尔新堡}}}，并在维也纳施塔莫斯多夫度过了他的童年。}
            & \cchar{什蒂普西奇出生于\mybox{red!15}{{{德国科恩堡}}}，在维也纳斯塔莫斯多夫度过了他的童年。} \\ \hline            
        Contradiction  & \cchar{纳舒厄白银骑士团队加入了夏季大学联盟，是本市的\mybox{red!15}{{{现役球队}}}。}
            & \cchar{Nashua Silver Knights 队是当前夏季联赛的一部分，也是该市的\mybox{red!15}{{{大学体育队}}}。} \\ \hline
         Entailment  & \ccharb{これらの見方は、\mybox{red!15}{{{福音主義的、清教徒的、プロ}}} \mybox{red!15}{{{テスタント的な}}}動きが出現するとともに、しばしば表明されてきました。}
            & \ccharb{これらの見解は多くの場合、\mybox{red!15}{{{新教徒}}}、清教徒、福音主義者が出現するなかで示されてきた。} \\ \hline            
         Contradiction  & \ccharb{1954 年\mybox{red!15}{{{にスリナムに戻った後、弁護士としてパ}}} \mybox{red!15}{{{ラマリボに}}}定住した。}
            & \ccharb{1954 年、\mybox{red!15}{{{パラマリボに戻ると、彼はスリナムで}}} \mybox{red!15}{{{弁護士 として定住しました}}}。} \\ \hline            
        \Xhline{3\arrayrulewidth}
    \end{tabular}}
    \caption{Case study on PAWS-X. mPMR can extract rationales to explain the sequence-pair classification in multiple languages.}
    \label{tab:clue_more}
\end{table*}

\section{Results and Analyses}
\paragraph{XLU Performance.}
Table~\ref{tab:XLU} shows the results on a variety of XLU tasks. mPMR outperforms all previous methods with an absolute improvement of 2.1 F1 over the best baseline (i.e. KMLM).
mPMR shows greater improvements over previous methods on span extraction tasks. 
In particular, mPMR achieves up to 7.3 and 7.1 F1 improvements over XLM-R on TyDiQA and WikiAnn respectively.
Such significant improvements probably come from the following two facts: (1) WikiAnn comprises a larger number of target languages (i.e. 40). Therefore, existing methods may struggle to align these low-resource languages with English due to a lack of language-specific data.
(2) TyDiQA is a more challenging cross-lingual EQA task with 2x less lexical overlap between the query and the answer than MLQA and XQuAD \cite{hu2020xtreme}.
Our mPMR, which acquires target-language span extraction capability from both MRC-style pre-training and English-only QA fine-tuning, achieves larger performance gains on more challenging task.


\paragraph{mPMR Pre-training.}
To reflect the impact of our MRC-style data and Stochastic Answer Position method on pre-training, we present a step-by-step analysis of the retrofitting process starting from XLM-R in Table~\ref{tab:pt}.
Our findings suggest that the significant improvements observed are largely due to the inclusion of multilingual MRC data.
Introducing English MRC data (model \#2) gives marginal improvements because model \#2 can only rely on cross-lingual representations to transfer the knowledge acquired during MRC-style pre-training.
When using MRC data on more languages (model \#4 and \#5), we can observe significant improvements on XLU tasks.
This can be attributed to the NLU capability directly inherited from MRC-style pre-training in target languages.
Additionally, with our Stochastic Answer Position method (model \#3), mPMR becomes more robust to position bias and thus improves XLU tasks.

\begin{figure}[t]
    \centering
    \includegraphics[scale=0.23]{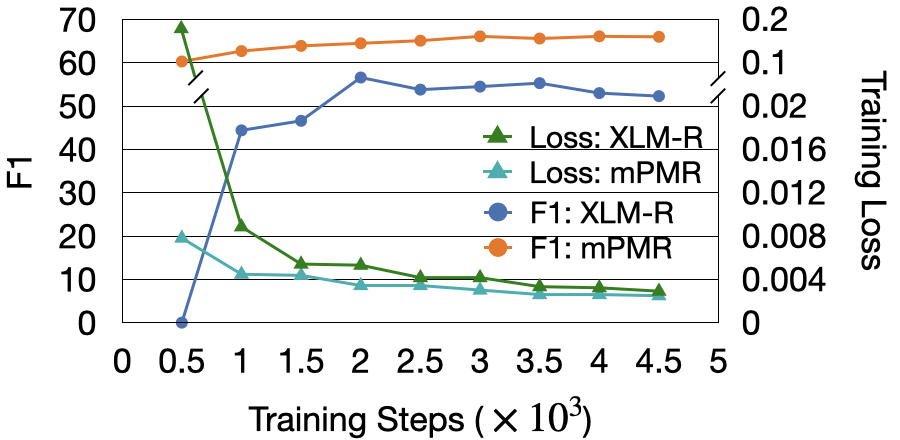}
    \caption{Convergence speed (Test set F1 and the training loss) of mPMR$_{\tt base}$ and XLM-R$_{\tt base}$ on WikiAnn. }
    \label{fig:ft}
\end{figure}

\paragraph{Explainable Sentence-pair Classification.}
Inspired by PMR \cite{xu2022clozing}, we investigated if the extraction capability of mPMR can be leveraged to explain sentence-pair classification.
Note that sentence-pair classification focuses on the inference between the two sentences. 
If we construct the query with only the task label as PMR does, such query does not solely correspond to any meaningful span in the context, and thus is hard to guide the span extraction.
Therefore, we leveraged another template ``{\tt [CLS]} \text{label}  \text{Sen-1} {\tt [SEP]} \text{Sen-2} {\tt [SEP]}'', where the two sentences are represented separately in the query and the context.
In this template, we can extract the exact span from \text{Sen-2} that leads to a contraction or entailment relation (i.e., the task label) with Sen-1.
Specifically, we passed the sentence pair to the model twice, with each sentence of the pair being designated as the Sen-2 respectively, and extract the context span with the highest probability score from both sentences.

As shown in Table~\ref{tab:clue_more}, the extracted spans are indeed important rationales that determine the relationship between two sentences.
Such a finding confirms that the extraction capability of mPMR can be appropriately used for explaining the sentence-pair classification process.
While the extraction capability may affect the learning of sequence classification during fine-tuning, resulting in a 0.4 Acc. decrease on XNLI.

\begin{figure}[t]
    \centering
    \includegraphics[scale=0.24]{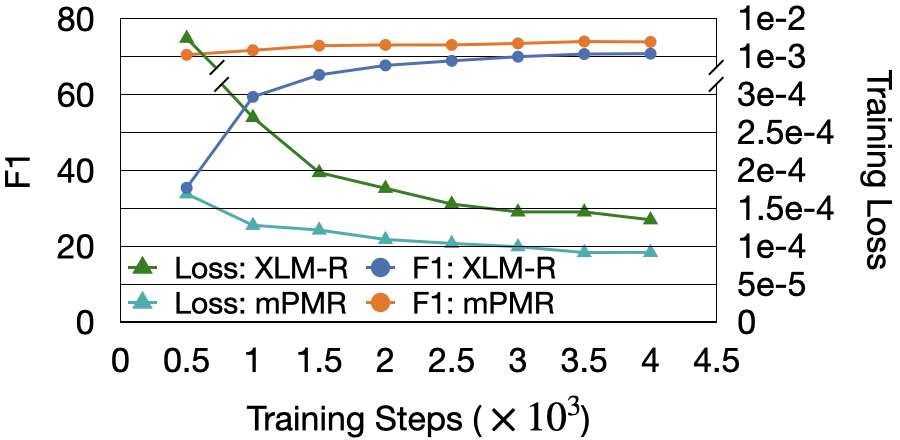}
    \caption{Convergence speed (Test set F1 and the training loss) of mPMR$_{\tt base}$ and XLM-R$_{\tt base}$ on XQuAD. }
    \label{fig:ft_xquad}
\end{figure}

\begin{figure}[t]
    \centering
    \includegraphics[scale=0.24]{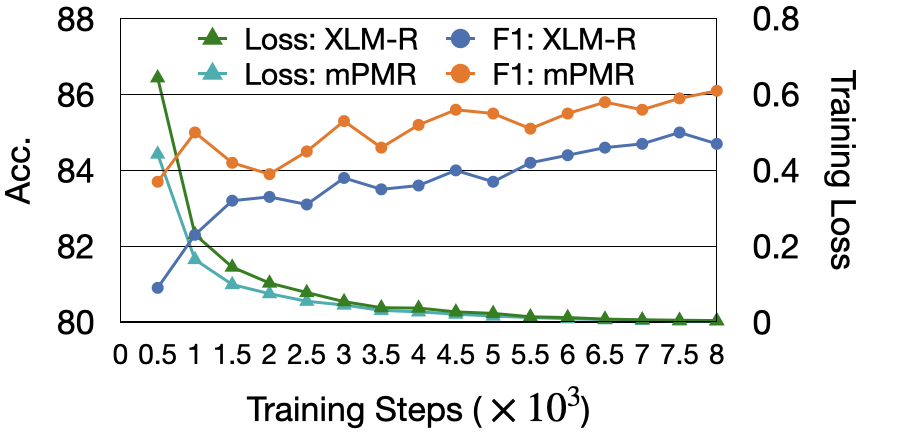}
    \caption{Convergence speed (Test set F1 and the training loss) of mPMR$_{\tt base}$ and XLM-R$_{\tt base}$ on PAWS-X. }
    \label{fig:ft_pawsx}
\end{figure}

\paragraph{mPMR Fine-tuning.}
We investigated the effects of mPMR on XLU fine-tuning. Figure~\ref{fig:ft} shows that mPMR converges faster than XLM-R on WikiAnn with an extremely low loss value even fine-tuned for 500 steps.
In terms of test set performance, mPMR outperforms XLM-R comprehensively and exhibits greater stability.
As a result, mPMR provides a better starting point for addressing XLU tasks compared to XLM-R. More examples from XQuAD and PAWS-X are provided in Figure~\ref{fig:ft_xquad} and \ref{fig:ft_pawsx}.

\section{Conclusions}
This paper presents a novel multilingual MRC-style pre-training method, namely mPMR.
mPMR provides a unified solver for cross-lingual span extraction and sequence classification and enables direct transfer of NLU capability from pre-training to downstream tasks.
mPMR clearly improves the previous baselines and provides a possible solution to explain the sentence-pair classification process.
\section*{Limitations}
We identify the following two limitations of our work:
\begin{itemize}
    \item Different from raw text, constructing MRC-style data from Wikipedia requires the existence of hyperlinks. This idea works well for resource-rich languages, such as English and Chinese. While such an idea is less effective for languages with few hyperlink annotations in Wikipedia because a small amount of MRC-style training data is difficult to guide the learning of NLU capability in those languages. A possible solution is to explore other data resources to  automatically construct large-scale MRC data for pre-training.
    \item As observed in Table~\ref{tab:XLU}, the improvements of sequence classification tasks are less significant than those of span extraction tasks. We suggest that the existence of anchors is not a strong relevance indicator between our constructed query and context. Such a finding is also observed in \citet{Chang2020Pre-training}. Therefore, constructing more relevant query-context pairs for sequence classification pre-training can possibly remedy this issue.
\end{itemize}

\bibliography{custom}
\bibliographystyle{acl_natbib}
\appendix

\clearpage
\section{Appendix}
\subsection{More Implementation Details}
\label{app:impl}
We collect the 2022-08-01 dump\footnote{https://dumps.wikimedia.org/enwiki/latest} of Wikipedia articles for the 24 languages in consideration.
The statistics of each language can be found in Table \ref{tab:statis}.
Then for each article, we extract the plain text with anchors via WikiExtractor \cite{Wikiextractor2015}.
Word tokenization is performed using spaCy\footnote{https://github.com/explosion/spaCy} if the language is supported, otherwise, we utilize PyThaiNLP\footnote{https://github.com/PyThaiNLP/pythainlp} for Thai and Sacremoses\footnote{https://github.com/alvations/sacremoses} for remaining languages.
For each anchor entity, we construct 10 answerable MRC examples and 10 unanswerable MRC examples as described in Sec.~\ref{sec:data}.
Anchor entities with low frequency (below 10 occurrences for English entities and 5 occurrences for entities in other languages) were excluded.

In mPMR, we use Huggingface's implementations of XLM-R~\cite{wolf-etal-2020-transformers}. 
During the pre-training stage, the query length $Q$ is set to 50 words, and the context length $C$ is set to 200 words. Both are computed before the subword segmentation.
We follow the default learning rate schedule and dropout settings used in XLM-R. We use AdamW~\cite{loshchilov2018decoupled} as our optimizer.
We train both mPMR$_{\tt base}$ and mPMR on 4 A100 GPU.
The learning rate is set to 1e-5, and the effective batch size for each step is set to 256 and 80 for mPMR$_{\tt base}$ and mPMR respectively in order to maximize the usage of the GPU memory.
We use the average scores of XQuAD, CoNLL, and PAWS-X to select the best mPMR checkpoint.
In fact, we continually pre-train mPMR$_{\tt base}$ and mPMR for 250,000 and 100,000 steps.
The training speed is around 6250 steps per hour.
The hyper-parameters of mPMR$_{\tt large}$ on downstream XLU tasks can be found in Table \ref{tab:full-params}.

\begin{table*}[]
    \centering
    \begin{tabular}{lrr|lrr}
    \toprule
         Language   & \# Entities   & \# MRC examples   & Language  & \# Entities   & \# MRC examples \\ \midrule
         ar         &  118,292      &  2,020,502        &  ko       & 94,616        & 1,597,076                  \\
         bn         &  25,081       &  410,634          &  nl       & 251,323       & 4,185,913                 \\
         de         &  864,746      &  14,795,826       &  pl       & 283,925       & 4,765,015                \\
         el         &  56,383       &  946,114          &  pt       & 216,695       & 3,648,603                  \\
         en         &  966,197      &  19,303,940       &  ru       & 432,437       & 7,342,472                  \\
         es         &  412,476      &  7,044,972        &  sv       & 169,030       & 2,808,214                  \\
         fi         &  113,118      &  1,960,636        &  sw       & 4,857         & 65,724                 \\
         fr         &  595,879      &  10,164,216       &  te       & 11,005        & 170,664                \\
         hi         &  15,350       &  242,078          &  th       & 31,676        & 522,434                \\
         id         &  70,960       &  1,164,662        &  tr       & 71,294        & 1,175,276                 \\
         it         &  376,417      &  6,421,850        &  vi       & 68,665        & 1,147,772                 \\
         ja         &  423,884      &  7,338,308        &  zh       & 259,785       & 4,438,004                 \\ \midrule
         &&& Total  &  5,934,091    & 103,680,905 \\ \bottomrule
    \end{tabular}
    \caption{Data statistics of mPMR pre-training data. The statistics is computed after removing the low-frequency entities. The number of MRC examples includes both answerable and unanswerable examples.}
    \label{tab:statis}
\end{table*}

\begin{table*}[]
    \centering
    \small
    \setlength{\tabcolsep}{1.5mm}{
    \begin{tabular}{@{}l||cccccccc@{}}\toprule
        Dataset         & \textbf{XQuAD}   & \textbf{MLQA}& \textbf{TyDiQA} & \textbf{WikiAnn} & \textbf{CoNLL}  & \textbf{SemEval16}  & \textbf{PAWS-X} & \textbf{XNLI} \\ \midrule
        Query Length    & 64        & 64    & 64    & 32    & 32    & 32   & 64   & 64   \\ 
        Input Length    & 384       & 384   & 384   & 192   & 192   & 192   & 192   & 192   \\
        Batch Size      & 8         & 8     & 8     & 16    & 16    & 32     & 16     & 32    \\
        Learning Rate   & 3e-5      & 3e-5  & 2e-5  & 1e-5  & 1e-5  & 2e-5  & 5e-5  & 3e-5  \\ 
        Epoch           & 3        & 3     & 10    & 10     & 10     & 20     & 10     & 3      \\\bottomrule
    \end{tabular}}
    \caption{Hyper-parameters settings in fine-tuning XLU tasks.}
    \label{tab:full-params}
\end{table*}

\subsection{Downstream XLU Tasks}
\label{app:dataset}
We evaluate mPMR on XLU tasks including both span extraction (EQA, NER, and ABSA) and sequence classification (sentence pair classification).
We follow \cite{xu2022clozing} to convert all tasks into MRC formulation and tackle them accordingly.
We show concrete examples for each task in Table \ref{tab:mrc_example}. 
Specifically, we evaluate the performance of EQA on three benchmarks: XQuAD \cite{artetxe-etal-2020-cross}, MLQA \cite{lewis-etal-2020-mlqa}, and TyDiQA \cite{clark-etal-2020-tydi} covering 11, 7, and 9 languages respectively.
For NER evaluation, we use the WikiAnn dataset \cite{pan-etal-2017-cross} restricted to the 40 languages from XTREME \cite{hu2020xtreme}, as well as the CoNLL dataset with 4 languages \cite{tjong-kim-sang-2002-introduction,tjong-kim-sang-de-meulder-2003-introduction}; We also evaluate the XLU performance of SemEval16 ABSA on 6 languages \cite{pontiki-etal-2016-semeval}, where we collect the data from \citet{li2020unsupervised,zhang-etal-2021-cross}.
Regarding the sequence classification task, we evaluate XNLI \cite{conneau-etal-2018-xnli} and PAWS-X \cite{yang-etal-2019-paws} with 15 and 7 languages respectively.

\begin{table*}[t]
    \centering
    \small
\renewcommand\arraystretch{1.5}
    \begin{tabular}{@{}p{18mm}||c|m{86mm}|m{37mm}@{}} 
    \toprule
         \textbf{Task}   && {\centering Example Input} & {\centering Example Output} \\
        \midrule
        \multirow{5}{*}{\makecell[c]{\textbf{EQA}\\(XSQuAD)}}  & \multirow{1}{*}{{\rotatebox[origin=c]{90}{{Ori.}}}}    & Question: Who lost to the Broncos in the divisional round? \;\;\;\;\;\;\;\;\;\;\;\;\;\;\;\; Context: The Broncos defeated the Pittsburgh Steelers in the divisional round, 23–16, by scoring 11 points in the final three minutes of the game. & Answer: "Pittsburgh Steelers"\\ \cline{2-4}
        & \multirow{1}{*}{\rotatebox[origin=c]{90}{{PMR}}}     &{\tt [CLS]} Who lost to the Broncos in the divisional round ? {\tt [SEP]} {\tt [SEP]} The Broncos defeated the Pittsburgh Steelers in the divisional round, 23–16 , by scoring 11 points in the final three minutes of the game . {\tt [SEP]} & (17,18) - "Pittsburgh Steelers" \\ 
        \midrule\midrule
        \multirow{13}{*}{\makecell[c]{\textbf{NER}\\(CoNLL)}}  & \multirow{1}{*}{{\rotatebox[origin=c]{90}{{Ori.}}}}    & Two goals in the last six minutes gave holders Japan an uninspiring 2-1 Asian Cup victory over Syria on Friday. & \makecell[l]{("Japan", LOC); \\ ("Syria", LOC); \\("Asian Cup", MISC)}\\ \cline{2-4}
        & \multirow{10}{*}{\rotatebox[origin=c]{90}{{PMR}}}     & {\tt [CLS]} "ORG" . Organization entities are limited to named corporate, governmental, or other organizational entities. {\tt [SEP]} {\tt [SEP]} Two goals in the last six minutes gave holders Japan an uninspiring 2-1 Asian Cup victory over Syria on Friday . {\tt [SEP]} & $\emptyset$ \\  \cline{3-4}
        && {\tt [CLS]} "PER" . Person entities are named persons or family . {\tt [SEP]} {\tt [SEP]} Two goals in the last six minutes gave holders Japan an uninspiring 2-1 Asian Cup victory over Syria on Friday . {\tt [SEP]} & $\emptyset$ \\ \cline{3-4}
        && {\tt [CLS]} "LOC" . Location entities are the name of politically or geographically defined locations such as cities , countries . {\tt [SEP]} {\tt [SEP]} Two goals in the last six minutes gave holders Japan an uninspiring 2-1 Asian Cup victory over Syria on Friday . {\tt [SEP]} & \makecell[l]{(32,32) - "Japan"; \\ (40,40) - "Syria"}   \\ \cline{3-4}
        && {\tt [CLS]} "MISC" . Examples of miscellaneous entities include events , nationalities , products and works of art . {\tt [SEP]} {\tt [SEP]} Two goals in the last six minutes gave holders Japan an uninspiring 2-1 Asian Cup victory over Syria on Friday . {\tt [SEP]} & (34,35) - "Asian Cup" \\ \midrule\midrule
        \multirow{7}{*}{\makecell[c]{\textbf{ABSA}\\(SemEval16)}}  & \multirow{1}{*}{{\rotatebox[origin=c]{90}{{Ori.}}}}    & Nice ambience, but highly overrated place. & \makecell[l]{("ambience", POS); \\ ("place", NEG)} \\ \cline{2-4}
        & \multirow{5}{*}{\rotatebox[origin=c]{90}{{PMR}}}     & {\tt [CLS]} "POS" . For aspect terms of positive sentiment . {\tt [SEP]} {\tt [SEP]} Nice ambience , but highly overrated place . {\tt [SEP]} & (13,13) - "ambience" \\  \cline{3-4}
        && {\tt [CLS]} "NEG" . For aspect terms of negative sentiment . {\tt [SEP]} {\tt [SEP]} Nice ambience , but highly overrated place . {\tt [SEP]} & (18,18) - "place"  \\ \cline{3-4}
        && {\tt [CLS]} "NEU" . For aspect terms of neutral sentiment . {\tt [SEP]} {\tt [SEP]} Nice ambience , but highly overrated place . {\tt [SEP]}  & $\emptyset$  \\ \midrule\midrule
        \multirow{8}{*}{\makecell[c]{\textbf{Sen. Pair}\\ \textbf{Classification}\\(PAWS-X)}}  & \multirow{1}{*}{{\rotatebox[origin=c]{90}{{Ori.}}}}    & \makecell[l]{Hypothesis: The Tabaci River is a tributary of the River Leurda in \\Romania.\\Premise: The Leurda River is a tributary of the River Tabaci in \\Romania.} & Contradiction\\ \cline{2-4}
        & \multirow{4}{*}{\rotatebox[origin=c]{90}{{PMR}}} & {\tt [CLS]} Contradiction . The hypothesis is a sentence with a contradictory meaning to the premise . {\tt [SEP]} {\tt [SEP]} Hypothesis : The Tabaci River is a tributary of the River Leurda in Romania . Premise : The Leurda River is a tributary of the River Tabaci in Romania . {\tt [SEP]} & (0,0) - "{\tt [CLS]}"  \\  \cline{3-4}
        && {\tt [CLS]} Entailment . The hypothesis is a sentence with a similar meaning as the premise . {\tt [SEP]} {\tt [SEP]} Hypothesis : The Tabaci River is a tributary of the River Leurda in Romania . Premise : The Leurda River is a tributary of the River Tabaci in Romania . {\tt [SEP]} & $\emptyset$  \\ \bottomrule
    \end{tabular}
    \caption{MRC examples of XLU tasks. We use English examples here for demonstration purposes. Ori. indicates the original data format of these tasks.}
    \label{tab:mrc_example}
\end{table*}



\subsection{mPMR Performance per Language}
We show the detailed results for each language in each task in Table \ref{tab:xquad_results} (XQuAD), Table \ref{tab:mlqa_results} (MLQA), Table \ref{tab:tydiqa_results} (TyDiQA), Table \ref{tab:wikiann_results} (WikiAnn), Table \ref{tab:conll_results} (CoNLL), Table \ref{tab:rest16_results} (SemEval16), Table \ref{tab:paws-x_results} (PAWS-X), and Table \ref{tab:xnli_results} (XNLI).

\begin{table*}[]
\resizebox{\textwidth}{!}{
\begin{tabular}{l|ccccccccccc|c}
\toprule
Model               & en        & ar        & de        & el        & es        & hi        & ru        & th        & tr        & vi        & zh            &  Avg.        \\
\midrule
XLM-R$_{\tt base}$  &82.2 / 72.0&65.5 / 49.9&73.9 / 59.7&71.2 / 56.3&76.3 / 59.4&66.4 / 52.0&73.7 / 58.9&64.7 / 54.6&67.0 / 52.8&73.3 / 54.7&65.0 / 55.9&70.8 / 56.9    \\
mPMR$_{\tt base}$   &84.4 / 73.4&69.6 / 53.2&76.4 / 61.5&74.9 / 58.4&77.4 / 60.2&69.2 / 54.5&75.2 / 58.8&69.2 / 57.6&70.4 / 55.8&74.8 / 55.8&71.8 / 65.5&74.0 / 59.5   \\
XLM-R               &86.5 / 75.6&72.4 / 54.8&79.3 / 63.0&79.2 / 61.6&82.0 / 62.9&76.1 / 59.1&79.0 / 62.9&72.2 / 59.8&75.4 / 60.8&79.7 / 60.8&68.2 / 58.2&77.3 / 61.7   \\
mPMR                &87.6 / 76.5&75.9 / 60.0&81.5 / 65.0&80.8 / 63.9&82.8 / 65.1&76.5 / 60.3&80.9 / 65.3&75.5 / 65.5&76.7 / 61.3&81.5 / 62.2&71.5 / 63.4&79.2 / 64.4   \\
\bottomrule
\end{tabular}}
\caption{XQuAD results (F1 / EM) for each language.}
\label{tab:xquad_results}
\end{table*}

\begin{table*}[]
\centering
\resizebox{\textwidth}{!}{
\begin{tabular}{l|ccccccc|c}
\toprule
Model               & en        & ar        & de        & es        & hi        & vi        & zh        & Avg.                  \\
\midrule
XLM-R$_{\tt base}$  &79.3 / 67.2&55.4 / 38.1&62.0 / 49.1&66.8 / 50.2&59.4 / 44.8&66.1 / 46.7&61.8 / 39.5&64.4 / 47.9    \\
mPMR$_{\tt base}$   &81.1 / 68.9&58.5 / 41.0&63.6 / 50.5&68.5 / 52.1&60.3 / 46.4&68.3 / 49.2&56.6 / 32.9&65.3 / 48.7   \\
XLM-R               &83.4 / 71.0&64.9 / 45.8&69.6 / 54.8&74.1 / 56.8&70.7 / 53.4&73.3 / 53.0&64.4 / 42.4&71.5 / 53.9    \\
mPMR                &84.0 / 71.4&66.4 / 47.0&70.3 / 56.2&74.5 / 57.1&71.4 / 54.1&74.7 / 54.4&70.5 / 47.3&73.1 / 55.4   \\

\bottomrule
\end{tabular}}
\caption{MLQA results (F1 / EM) for each language.}
\label{tab:mlqa_results}
\end{table*}

\begin{table*}[]
\centering
\resizebox{\textwidth}{!}{
\begin{tabular}{lccccccccc|c}
\toprule
Model               & en        & ar        & bn        & fi        & id        & ko        & ru        & sw        & te        & Avg. \\
\midrule
XLM-R$_{\tt base}$  &66.8 / 57.3&55.7 / 42.0&31.5 / 20.4&52.6 / 40.3&69.1 / 55.6&36.3 / 27.9&54.8 / 36.5&53.0 / 34.7&37.4 / 28.8&50.8 / 38.2\\
mPMR$_{\tt base}$   &71.1 / 61.6&66.3 / 52.6&56.5 / 41.6&65.5 / 53.1&73.9 / 63.7&50.4 / 38.8&64.4 / 37.9&57.4 / 41.1&65.3 / 50.4&63.4 / 49.0   \\
XLM-R               &71.3 / 60.7&69.3 / 52.3&66.2 / 53.1&64.3 / 51.3&76.5 / 62.5&58.3 / 46.7&64.7 / 43.4&68.6 / 53.1&67.3 / 41.1&67.4 / 51.6   \\
mPMR                &76.4 / 65.2&76.0 / 58.0&72.3 / 55.8&74.4 / 56.5&84.1 / 71.3&62.2 / 50.7&72.5 / 43.2&76.5 / 63.1&77.7 / 60.8&74.7 / 58.3   \\
\bottomrule
\end{tabular}
}
\caption{TyDiQA-GoldP results (F1 / EM) for each language.}
\label{tab:tydiqa_results}
\end{table*}

\begin{table*}[]
\centering
\resizebox{\textwidth}{!}{
\begin{tabular}{l|cccccccccccccccccccc}
\toprule
Model               & en    & af    & ar    & bg    & bn    & de    & el    & es    & et    & eu    & fa    & fi    & fr    & he    & hi    & hu    & id    & it    & ja    & jv     \\
\midrule
XLM-R$_{\tt base}$  & 84.2  & 75.3  & 47.3  & 79.0  & 66.3  & 77.5  & 75.3  & 78.0  & 69.6  & 56.0  & 38.1  & 70.4  &81.4   &50.8   &67.9   & 72.4  &51.0   &79.6   &19.6   & 63.9  \\
mPMR$_{\tt base}$   & 85.1  & 80.7  & 57.6  & 80.2  & 71.9  & 81.2  & 77.6  & 79.5  & 79.1  &71.3   & 49.6  & 80.4  & 82.4  & 65.2  & 71.7  & 82.2  & 58.6  & 83.5  & 43.2  & 72.0  \\
XLM-R               & 85.4  &81.1   &53.9   &84.0   &73.8   &82.3   &82.8   &80.4   &68.8   &54.8   &64.2   &75.9   &81.4   &59.3   &72.9   &76.4   & 59.3  & 84.6  & 13.2  & 71.2  \\
mPMR                & 86.0  & 81.7  & 56.1  & 85.9  & 79.6  & 82.3  & 82.3  & 75.5  & 82.7  & 69.6  & 75.2  & 84.1  & 82.0  & 66.5  & 75.9  & 84.0  & 59.9  & 86.1  & 49.1  & 72.4  \\
\midrule
                    & ka    & kk    & ko    & ml    & mr    & ms    & my    & nl    & pt    & ru    & sw    & ta    & te    & th    & tl    & tr    & ur    & vi    & yo    & zh            \\
\midrule
XLM-R$_{\tt base}$  & 58.7  &40.6   & 34.3  &50.8   &46.0   &63.8   &40.6   &81.5   &80.0   & 65.4  &76.1   &43.0   &46.4   &4.2    &71.9   &68.7   &45.7   &70.9   & 1.5   &23.0  \\
mPMR$_{\tt base}$   & 72.2  & 45.1  & 52.9  & 62.4  & 59.4  & 68.1  & 57.4  & 83.7  &81.5   &71.8   & 77.3  & 50.5  & 57.4  & 3.0   & 74.2  & 80.3  & 55.7  & 75.2  &31.6   & 49.9  \\
XLM-R               & 59.9  & 41.7  & 41.3  &56.8   & 58.2  & 76.7  & 29.6  &86.1   &85.2   &72.2   &77.6   & 52.3  & 51.6  & 7.1   &78.8   & 70.9  & 64.0  &80.0   &27.2   &22.4  \\
mPMR                & 77.3  & 46.8  & 57.9  & 70.6  &68.1   &73.8   &57.8   & 86.0  &83.6   & 72.8  &79.8   & 62.6  & 58.1  & 3.8   & 83.0  & 80.3  &76.2   &83.6   &36.1   &54.4 \\
\bottomrule
\end{tabular}}
\caption{WikiAnn results (F1 Score) for each language.}
\label{tab:wikiann_results}
\end{table*}

\begin{table*}[]
\centering
\begin{tabular}{l|ccccccc|c}
\toprule
Model               &en     & de    & es    & nl    & Avg.  \\
\midrule
XLM-R$_{\tt base}$  & 91.3  & 71.0  & 78.7  & 75.7  &79.2  \\
mPMR$_{\tt base}$   & 91.9  & 74.3  & 80.8  & 79.7  & 81.7  \\
XLM-R               & 92.8  & 73.7  & 81.6  & 77.7  & 81.4  \\
mPMR                & 93.5  & 75.0  & 85.0  & 83.1  & 84.1  \\
\bottomrule
\end{tabular}
\caption{CoNLL results (F1 Score) for each language.}
\label{tab:conll_results}
\end{table*}

\begin{table*}[]

\centering
\begin{tabular}{l|cccccc|c}
\toprule
Model               &en     & es    & fr    & nl    & ru    & tr    & Avg.  \\
\midrule
XLM-R$_{\tt base}$  &76.5   & 65.4  &55.6   &61.2   &56.1   &45.4   &60.0      \\
mPMR$_{\tt base}$   &77.6   &68.6   &56.4   &62.2   &59.5   &48.4   &62.1   \\
XLM-R               &82.4   &71.3   &60.3   &67.4   &61.2   &49.1   &66.1   \\
mPMR                &82.8   &71.9   &64.7   &67.4   &66.9   &55.7   &68.2   \\
\bottomrule
\end{tabular}
\caption{SemEval16 results (F1 Score) for each language.}
\label{tab:rest16_results}
\end{table*}

\begin{table*}[]
\centering
\begin{tabular}{l|ccccccc|c}
\toprule
Model               &en     & de    & es    & fr    & ja    & ko    & zh    & Avg.  \\
\midrule
XLM-R$_{\tt base}$  & 94.3  & 87.7  & 89.1  & 88.7  & 77.0  & 76.6  & 81.3  & 85.0  \\
mPMR$_{\tt base}$   & 94.3  & 88.4  & 90.1  & 88.9  & 79.0  & 79.4  & 82.4  & 86.1  \\
XLM-R               & 95.2  & 89.3  & 91.0  & 90.9  & 79.6  & 79.9  & 82.5  & 86.9  \\
mPMR                & 95.2  & 90.6  & 90.3  & 91.3  & 81.2  & 82.9  & 84.6  & 88.0  \\
\bottomrule
\end{tabular}
\caption{PAWS-X accuracy scores (Acc.) for each language.}
\label{tab:paws-x_results}
\end{table*}

\begin{table*}[]
\resizebox{\textwidth}{!}{
\begin{tabular}{l|ccccccccccccccc|c}
\toprule
Model               & en    & ar    & bg    & de    & el    & es    & fr    & hi    & ru    & sw    & th    & tr    & ur    & vi    & zh    & Avg.  \\
\midrule
XLM-R$_{\tt base}$  & 84.6  & 71.0  & 76.8  & 75.6  & 74.9  & 77.9  & 76.9  & 68.9  & 74.1  & 64.4  & 71.1  & 72.4  & 65.2  & 73.2  & 73.0  & 73.3 \\
mPMR$_{\tt base}$   & 84.2  & 71.5  & 77.2  & 75.5  & 75.5  & 78.6  & 76.9  & 69.5  & 74.7  & 62.5  & 71.4  & 71.6  & 65.5  & 74.3  & 74.0  & 73.6 \\
XLM-R               & 88.2  & 77.0  & 81.7  & 81.2  & 81.2  & 84.2  & 81.7  & 74.9  & 78.9  & 70.8  & 75.7  & 77.4  & 70.6  & 78.0  & 77.7  & 78.6 \\
mPMR                & 88.3  & 77.9  & 82.9  & 82.2  & 81.0  & 83.5  & 82.2  & 75.2  & 79.8  & 71.2  & 76.1  & 78.9  & 71.6  & 78.9  & 79.0  & 79.3 \\
\bottomrule
\end{tabular}}
\caption{XNLI accuracy scores (Acc.) for each language.}
\label{tab:xnli_results}
\end{table*}

\end{document}